\documentclass{article}

    \PassOptionsToPackage{numbers, compress}{natbib}

    \usepackage[final]{neurips_2023}

\usepackage[utf8]{inputenc} 
\usepackage[T1]{fontenc}    
\usepackage{hyperref}       
\usepackage{url}            
\usepackage{booktabs}       
\usepackage{amsfonts}       
\usepackage{nicefrac}       
\usepackage{microtype}      
\usepackage{xcolor}         
\usepackage{graphicx}
\usepackage{alphalph}
\usepackage{tabularx}
\usepackage{array}
\newcolumntype{Y}{>{\centering\arraybackslash}X}
\makeatletter
\newcolumntype{|}{!{\vrule \@width \arrayrulewidth}}
\makeatother
\usepackage{algorithm}
\usepackage[noend]{algpseudocode}
\usepackage{wrapfig}

\title{Denoising Heat-inspired Diffusion with Insulators for Collision Free Motion Planning }

\author{%
  Junwoo~Chang$^{*1}$,\ \ Hyunwoo~Ryu$^{*1}$,\ \ Jiwoo~Kim$^{1}$,\ \ Soochul~Yoo$^{1}$ \ \textbf{Jongeun~Choi$^{\dagger 1,2}$}\ \\  \textbf{Joohwan~Seo$^{2}$}\textbf{,}\ \ \textbf{Nikhil~Prakash$^{2}$}\textbf{,}\ \ \textbf{Roberto~Horowitz$^{2}$} \\
  $^{1}$Yonsei University,\ \ $^{2}$University of California, Berkeley\\  
  \texttt{\{junwoochang,tomato1mule,nfsshift9801,usam205,jongeunchoi\}@yonsei.ac.kr}\\
  \texttt{\{joohwan\_seo, nikhilps, horowitz\}@berkeley.edu}\\
  $^*$Contributed Equally. $\quad^\dagger$Corresponding author.
}

\begin{document}

\maketitle

\begin{abstract}
  Diffusion models have risen as a powerful tool in robotics due to their flexibility and multi-modality. While some of these methods effectively address complex problems, they often depend heavily on inference-time obstacle detection and require additional equipment. Addressing these challenges, we present a method that, during inference time, simultaneously generates only reachable goals and plans motions that avoid obstacles, all from a single visual input. Central to our approach is the novel use of a \emph{collision-avoiding diffusion kernel} for training. Through evaluations against behavior-cloning and classical diffusion models, our framework has proven its robustness. It is particularly effective in multi-modal environments, navigating toward goals and avoiding unreachable ones blocked by obstacles, while ensuring collision avoidance. Project Website: \url{https://sites.google.com/view/denoising-heat-inspired}
\end{abstract}

\section{Introduction}
Diffusion probabilistic models have been recognized for multi-modality in various fields \cite{song2023loss,meng2023distillation,kawar2023imagic,de2022convergence,rombach2022high,gao2020learning,somepalli2023understanding,liu2023unsupervised,chung2023parallel,trippe2022diffusion,yim2023se}. While motion planning \cite{kavraki1996probabilistic,lavalle1998rapidly,ni2022ntfields,ni2023progressive} and manipulation \cite{seo2023robot,kim2023robotic,ryu2022equivariant,simeonov2022neural} often present complex challenges, these diffusion models provide effective solutions \cite{janner2022planning,xiao2023safediffuser,pearce2023imitating,pearce2022counter,carvalho2022conditioned,carvalho2023motion,pearce2022counter,ryu2023diffusion,urain2023se}, highlighting their growing significance in robotics. 
In particular, \citet{urain2023se} introduced a novel method that utilizes guided diffusion to generate both goals and motions. Compared to the process where goals are generated and motions are planned separately, this integrated approach is more efficient and has the strength of generating only reachable goals. However, this approach necessitates comprehensive knowledge of obstacles, potentially requiring equipment such as LIDAR during the inference time, which can be costly and time-consuming.

To overcome this issue, we introduce an alternative approach that utilizes a diffusion model with only a single visual input. Rather than relying on diffusion guidance to direct collision avoidance, our method incorporates this behavior into the motion planning process. The core of our approach is leveraging a diffusion model with a collision-avoiding diffusion kernel. Drawing inspiration from the behavior in which heat avoids insulators, similar to collision avoidance motions, 
we have adapted the modified heat equation to build our diffusion kernel that mimic behaviors of heat dispersion with insulating obstacles.
As a consequence, our approach operates efficiently on just a top-down view image during the inference, making it well-suited for real-world scenarios. 

\begin{figure}
  \centering
  \includegraphics[width=1.0\linewidth]{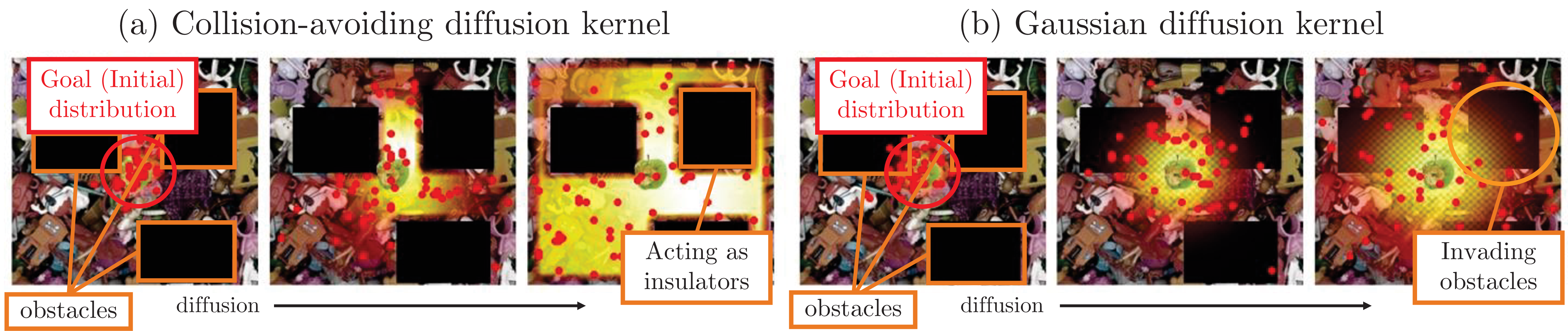} \vspace{-0.4cm}
  \caption{\textbf{Collision avoiding diffusion kernel.} The comparison clearly shows the differences between two diffusion kernels in an environment with obstacles.  \textbf{(a)} The collision-avoiding diffusion kernel moves without invading any obstacles. \textbf{(b)} In contrast, the Gaussian diffusion kernel often runs into obstacles, indicating a higher risk of collisions.}
  \label{dist}
\end{figure}

We summarize our contributions as: (1) We introduce an approach leveraging collision-avoiding diffusion kernel within the diffusion model, presenting a clear distinction from Gaussian distributions, as seen in Fig.\ref{dist}.
(2) Our method offers an end-to-end motion planning technique that, using only a top-down view image as input during inference, ensures collision avoidance and generates only reachable goals. This approach eliminates the need for detailed object state information and potential auxiliary equipment, making the process of inferring obstacle avoidance more efficient.
Our experiments highlight the model's capability to generate reachable goals while avoiding collisions and exhibiting multi-modality against baseline methods.
\section{Background}
\label{bg}
\paragraph{Problem statement.}
Let's define the state space in a $2D$ environment, $\mathcal{X}\subset \mathbb{R}^2$. The subset of $\mathcal{X}$ where obstacles exist is $\mathcal{X}_{obs}\subset\mathcal{X}$. The feasible space is $\mathcal{X}_{free}=\mathcal{X}\backslash \mathcal{X}_{obs}$. The primary objective is to generate the goal as well as the state sequence $\tau= \{x_1,...,x_T\}$ where each $x_t\in\mathcal{X}_{free}$. Given only a top-down view map image, $y$, our method aims to generate the sequence $\tau$ that reaches the goal distribution, $p_g(x|y)$. We solve the problem by leveraging the concepts of the diffusion model.
\paragraph{Score-based model learning and sampling.}
Diffusion models work by repeatedly denoising a noisy distribution \cite{ho2020denoising,song2019generative}. Such models are able to enable the generation of sequence $\tau$ using a visual input. We can express $p_t(x|y)$ as the distribution at a diffusion step, conditioned on the given image, and $p_0(x|y)$ as the distribution centered around the goal state with minimal variance, equivalent to $p_g(x|y)$. Using anealed Langevin dynamics, we can generate the sequence $\tau$, provided $p_t(x|y)$ excluding any $x\in\mathcal{X}_{obs}$. Training involves adding noise to the goal distribution using a diffusion kernel, denoted as $p_{0t}(x_t|x_0,y)$. This process gradually transforms the original distribution into a noised one given by $p_t(x_t|y) = \int p_{0t} (x_t|x_0,y)p_0(x_0|y)dx_0$. The model's goal is to accurately estimate the score $s_{\theta^*}(x_t)=\nabla_{x_t}\log p_t(x_t|y)$, similarly to \citet{welling2011bayesian}.
After training, the model can generate states from the perturbed distribution using annealed Langevin dynamics sampling \cite{song2019generative}. With a sufficiently small stepsize and repeated samplings, it is proven that the ultimately generated $\tilde{x}$ belongs to the goal distribution $p_0(x|y)$ \cite{song2019generative}.

\section{Goal generation with collision-avoiding diffusion kernel}
\paragraph{Collision avoiding diffusion kernel.}
To develop a diffusion kernel that provides collision avoidance as detailed in Section \ref{bg}, we derive from the heat equation:
\begin{equation}
    \frac{\partial u}{\partial t} = \nabla\cdot\left (K(x)\nabla u \right )
    \label{heateq}
\end{equation}
\begin{figure}
  \centering
  \includegraphics[width=.9\linewidth]{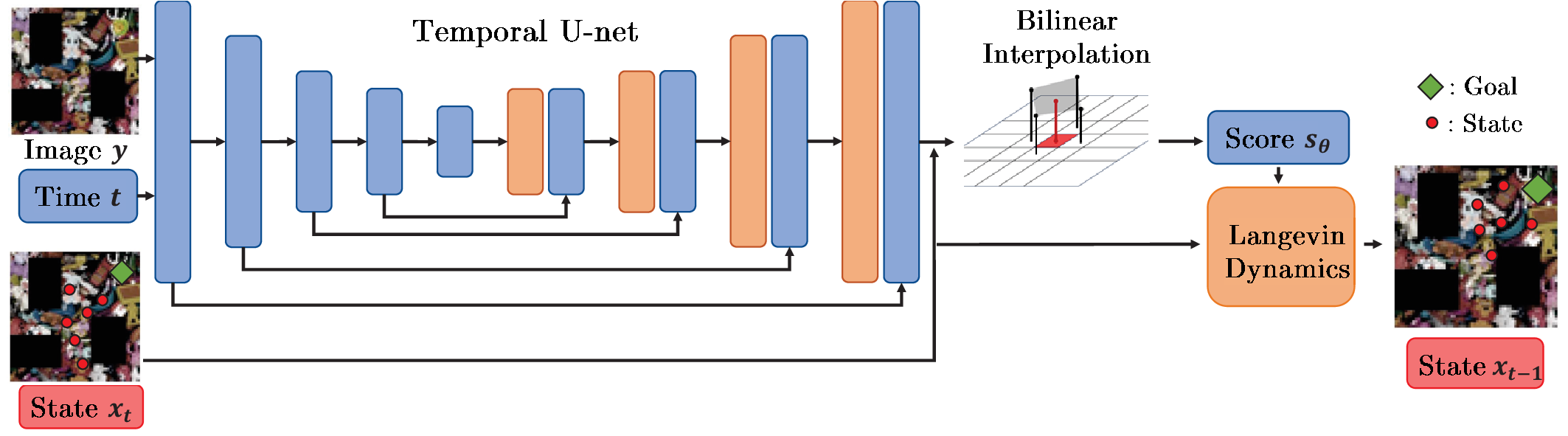} 
  \vspace{-0.2cm}
  \caption{\textbf{Architecture and overview of our method.} The model processes the visual input $y$ and time $t$ to produce an output score field, which is then bilinearly interpolated with the input state $x_t$ to obtain the score value at $x_t$. To determine the next state leading to the goal, we employ annealed Langevin dynamics sampling.}
  \label{arch}
\end{figure}
However, we employed a modified version, detailed in Appendix \ref{revisedheat}, selected for its computational efficiency and compatibility with our model. The core of our approach is the collision-avoiding diffusion kernel $p_{0t}(x|y)$, which is based on the heat conduction principles. The model estimates the non-conductive areas from the map image $y$, identifying obstacles and map boundaries as zero-conductivity regions. To compute this kernel, we solve the modified heat equation, where the heat distribution $u$ at any given time step is regarded as the distribution $p_{0t}(x|y)$. This distribution evolves from the goal distribution $p_0(x|y)$ while avoiding non-conductive regions, effectively preventing collisions as the robot navigates toward the goal by the reverse diffusion process.
\paragraph{Training with the collision avoiding diffusion kernel.}
Target scores for our model are derived by resolving the heat equation with obstacle consideration to form a collision-avoiding diffusion kernel. This field is normalized to probability, logarithmically adjusted, and then differentiated to yield the score field. The desired score for a sampled state $x_t$ is then directly obtained from this field. Comprehensive methods are detailed in Appendix.\ref{targetscore}.  The training objective is formalized as:
\begin{equation}
    L(\theta)=\mathbb{E}_t  \left (\lambda(t)\mathbb{E}_{x_0}\mathbb{E}_{x_t|x_0,y_{obs}}[\|s_\theta(x_t,t,y)-\nabla_{x_t}\log p_{0t}(x_t|x_0,y_{obs})\|^2_2] \right )
    \label{eq1}
\end{equation}
where $y_{obs}$ represents the obstacle mask, and $\lambda(t)=f(\sigma_t)^2$. Here, $\sigma_t=\sqrt{2k}$ is the standard deviation (std) of the Gaussian heat kernel. The function $f$ stabilizes $\sigma_t$ once it equals half the image size, approximating the std of the distribution confined by the map boundaries. We use the U-net structure for the model from \citet{ho2020denoising}.
\paragraph{Generating states for motion planning.}
The overall sequence of the inference is described in Fig.\ref{arch}. Given a randomly generated map image and an initial state $x_T$, we iteratively produce a sequence of states $x_t\in\mathcal{X}_{free}$. This generation is achieved by addressing the reverse SDE with our trained score model, leveraging the annealed Langevin dynamics sampling (details in Appendix \ref{ALD}):
\begin{equation}
    \tilde{x}_{t-1}=\tilde{x}_t+\frac{\alpha_t}{2}s_\theta(x_t, t, y)+\sqrt{\alpha_t} z_t,~~~~z_t\sim\mathcal{N}(0,I)
\end{equation}
where $\alpha_t=\epsilon\cdot\frac{\lambda(t)}{\lambda(T)}$. In practice, we set parameters as $\epsilon=0.0008,~T=10$, and performed 100 annealing iterations. Every generated state belongs to $p_t(x|y)$ \cite{song2019generative}, progressively approaching the goal distribution, $p_0(x|y)$, with collision avoidance.
\begin{figure}[h]
  \centering  \includegraphics[width=.9\linewidth]{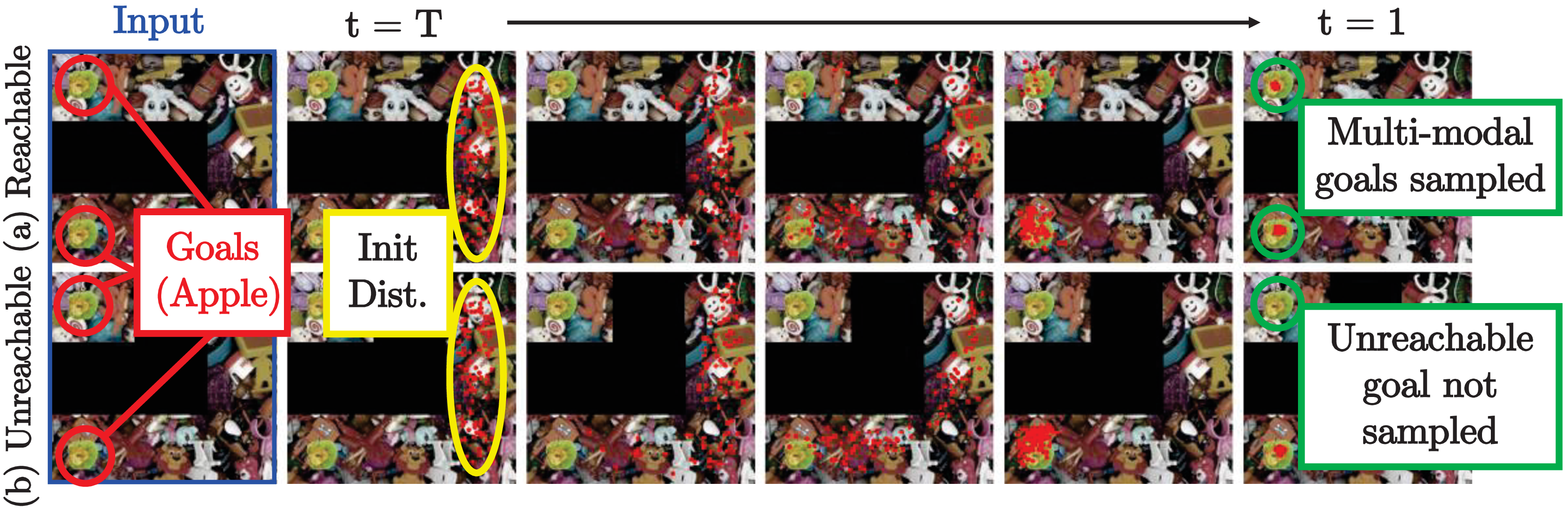} \vspace{-0.3cm}
  \caption{\textbf{Experiment results of our method.} In the $64\times 64$ input image, black areas indicate obstacles, red dots illustrate states originating from the initial distribution, and green apples mark the goals. \textbf{(a)} The first row demonstrates an experiment with two multi-modal goals generated. It proves the multi-modality of our model by moving the states toward each goal. \textbf{(b)} The second row presents a similar setup, but with one goal being unreachable. It only generates the reachable goal without any collisions with the obstacle.}
  \label{exp1}
\end{figure}
\section{Experiments}
We assessed our method's performance using a benchmark experiment on a noisy map from \citet{pearce2023imitating}, setting it against three established baseline models for comparison. The first model is expert-data behavior cloned (BC). The second and third, named Gaussian diffusion and Gaussian diffusion + RRT*, utilize a Vanilla diffusion model for goal generation, with the latter also incorporating RRT* for planning. 
\begin{table}[ht]
  \centering
  \footnotesize
  \setlength{\tabcolsep}{3pt} 
  \caption{\textbf{Experiment results}. Success rates and KL divergence for $10$K samples, aggregated from $100$ samples across $100$ episodes. Whenever samples encountered obstacles or if there was a failure in trajectory generation, they remained stationary at their current location.}
  \begin{tabularx}{\textwidth}{lYYYYYY}
    \toprule
    & \multicolumn{2}{c}{Uni-modality} & \multicolumn{2}{c}{Multi-modality} & \multicolumn{2}{c}{With unreachable goal} \\ 
    \cmidrule(lr){2-3} \cmidrule(lr){4-5} \cmidrule(lr){6-7}
    Models & Success rate & KL diverg. & Success rate & KL diverg. & Success rate & KL diverg. \\
    \midrule
    BC & \textbf{100}. & 6.44 & 0. & 13.32 & 0. & 14.83 \\
    Gaussian diffusion & 34.23 & 10.37 & 57.52 & 11.91 & 30.23 & 12.32 \\
    Gaussian diff + RRT* & 98.81 & 6.36 & 97.31 & \textbf{6.34} & 50.31 & 23.01 \\
    \textbf{Ours} & 98.5 & \textbf{6.05} & \textbf{99.98} & 6.45 & \textbf{100.} & \textbf{6.46} \\ 
    \bottomrule
  \end{tabularx}
  \label{tble}
\end{table}
All models under comparison employ the same U-net architecture and are described in Appendix \ref{pseudo}. Our evaluation criteria encompass Uni-modality, Multi-modality, and With unreachable goal generation. Uni-modality measures performance in reaching a single goal, Multi-modality assesses success in multi-goal scenarios, and With unreachable goal generation evaluates the model's ability to avoid generating unreachable goals. During the initial experiment, a single goal exists, and two goals in the next experiment. In the final experiment, an obstacle is added near one of the goals, making it unreachable.
\begin{wrapfigure}{r}{0.4\textwidth}
  \centering
  \includegraphics[width=1.\linewidth]{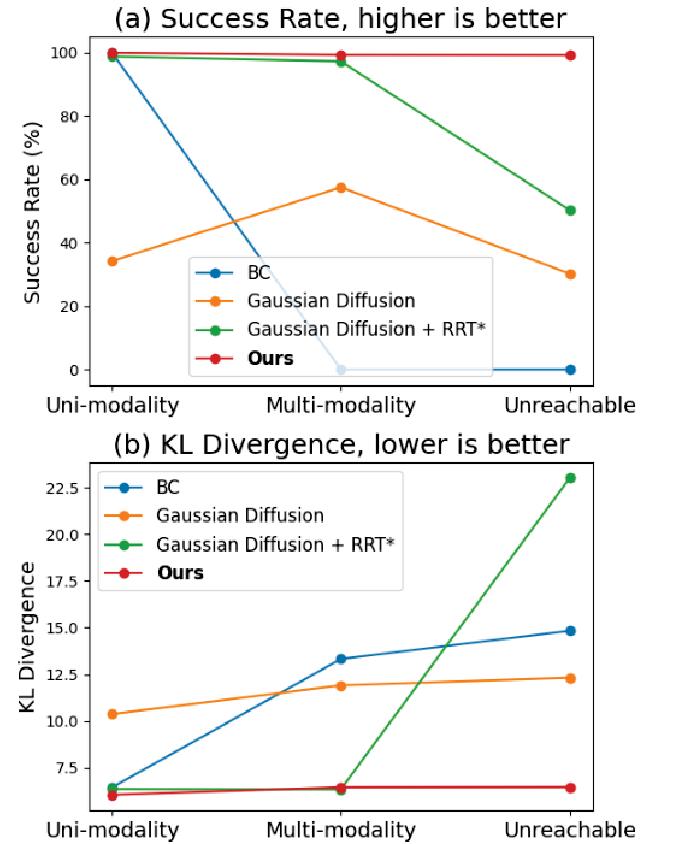}
  \caption{\textbf{Graphical analysis of experiment results}. \textbf{(a)}: success rate of reaching goal distributions. \textbf{(b)} KL divergence between the goal distributions and the empirical distribution. Both indicates that our method shows robust and good performance in all scenarios, compared to baselines.}
  \label{graph}
  \vspace{-40pt}
\end{wrapfigure}
\paragraph{Analysis.} In Table \ref{tble}, we detail the success rate, the count of states that successfully reach any goal distributions, and the KL divergence, a metric measuring the divergence between the empirical distribution of generated samples and ground-truth goal distribution. 
The BC model performed well in a single-goal environment. However, it lacked multi-modality, as a consequence of its limitations from training with MSE loss, as noted in \citet{pearce2023imitating}. The Gaussian diffusion lagged in both goal attainment and divergence, as it doesn't account for collision avoidance. 
The Gaussian diffusion + RRT* performed well in both the Uni-modality and Multi-modality tests. Yet, it mistakenly generated the unreachable goal in the last experiment, performing about $50\%$ success rate. In contrast, our method consistently showcased strong performance across all experiments. As illustrated in Fig.\ref{graph}, our method consistently maintains high success rate and low KL divergence, indicating a reliable performance in achieving the desired goal distributions. The results highlight our method's potential performance in real-world tasks. The visual results are displayed in Fig.\ref{exp1} and Appendix \ref{addexp}.
\section{Conclusion}
We have introduced an end-to-end method with a conditioned diffusion model to generate reachable goals without collisions with just a visual input. The core of our method is the utilization of a heat-inspired collision-avoiding diffusion kernel within a diffusion model to ensure collision avoidance. 
Furthermore, it can effectively avoid generating unreachable goals.
Thus, we presented a framework that is user-friendly in the sense that it does not require inference time collision checking or any auxiliary equipment. Future work involves extending this approach to accommodate high-dimensional sensory inputs in the field of robotics.

\begin{ack}
This work was supported by the National Research Foundation of Korea (NRF) grants funded by the Korea government (MSIT) (No.RS-2023-00221762 and No. 2021R1A2B5B01002620). This work was also supported by the Korea Institute of Science and Technology (KIST) intramural grants (2E31570), and a Berkeley Fellowship.
\end{ack}

\small
\bibliography{neurips_2023}
\bibliographystyle{unsrtnat}

\newpage
\section*{Appendix}
\appendix
\section{Benchmark experiment results}
\label{addexp}
\begin{figure}[ht]
  \centering
  \includegraphics[width=1.\linewidth]{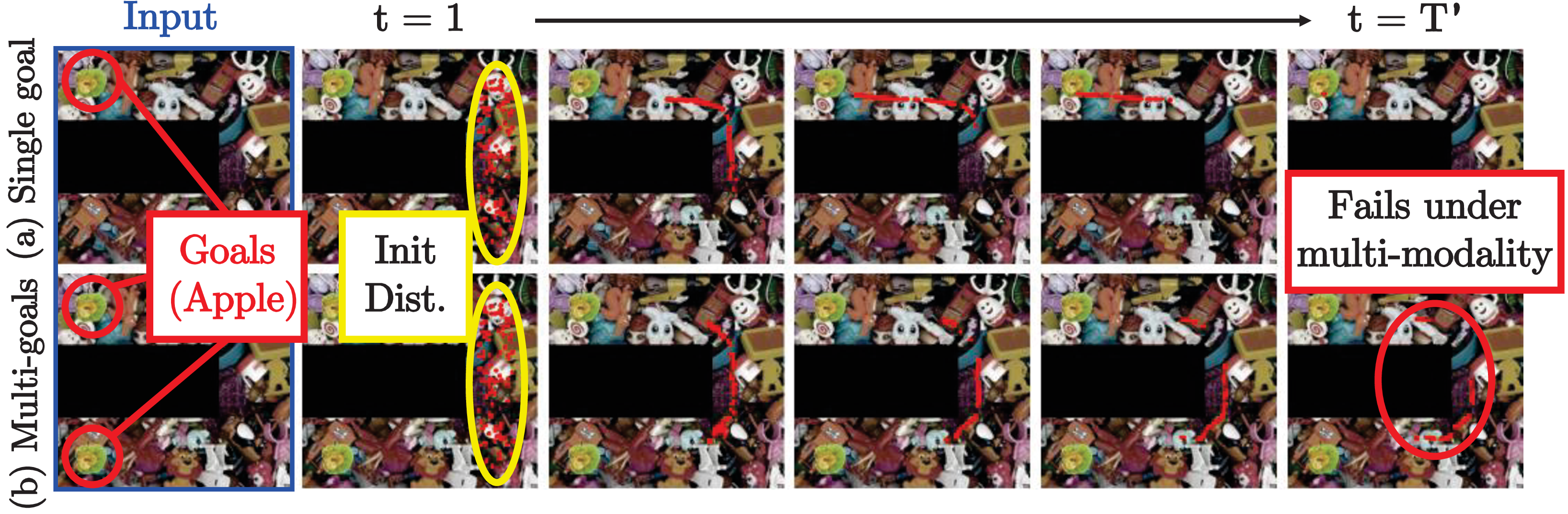}
  \caption{\textbf{Performance evaluation of the BC model in terms of multi-modality.} Black areas denote obstacles, red dots represent states initiated from the initial distribution, and green apples mark the goals. \textbf{(a)} The first row presents a scenario with one reachable goal generated. \textbf{(b)} The second row displays a test similar to the first but with multi-modal goals generated. It indicates that the behavior-cloning method performs well in uni-modal scenarios, but lacks the multi-modality.}
\end{figure}
\begin{figure}[ht]
  \centering
  \includegraphics[width=1.\linewidth]{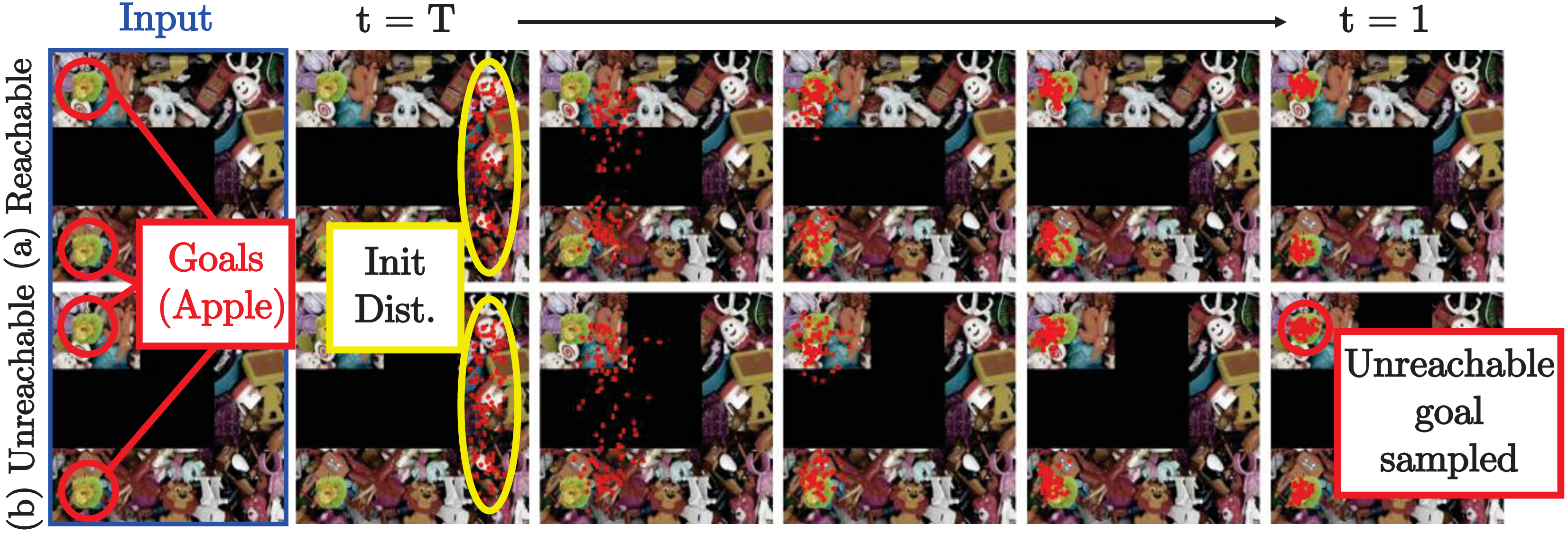}
  \caption{\textbf{Performance evaluation of the Gaussian model in sampling reachable goals.} In the visual representation, black areas denote obstacles, red dots illustrate states generated by the Gaussian model from the initial distribution, and green apples mark the goals. \textbf{(a)} The first row demonstrates an experiment with two reachable multi-modal goals generated. \textbf{(b)} The second row presents a scenario similar to the first, but with one unreachable goal. We use this method directly for planning or add another motion planner to move the states toward the generated goals. The result proves that the Gaussian diffusion model performs the multi-modality, but cannot assure reachable goal generation or collision avoidance.}
\end{figure}
\section{Detailed sampling implementation}
\label{ALD}
To address convergence speed issues, \citet{ryu2023diffusion} modifies the original Langevin dynamics sampling equation as follows:
\begin{equation}
    \tilde{x}_{t-1} = \tilde{x}_{t} + \frac{\epsilon}{2}s_\theta(x_t, t)t^{k_1}+t^{\frac{k_1+k_2}{2}}z_t,~~~~z_t\sim\mathcal{N}(0,I)
\end{equation}
This approach exhibits impressive performance in control tasks, particularly in state sampling without collisions, mirroring our scenario. While they modify $\alpha_t$ to $t^{k_1}$ for their purposes, we opt for $\alpha_t^{k_1}$ for clarity, making the \textit{temperature} term adapt as $\alpha_t^{k_2}$. Given that in our configuration $\alpha_t=\epsilon\cdot\frac{\lambda(t)}{\lambda(T)}$ and $\alpha_t\rightarrow 0$ as $t\rightarrow 0$, our adapted Langevin dynamics sampling equation can be expressed as:
\begin{equation}
    \tilde{x}_{t-1} = \tilde{x}_{t} + \frac{\alpha_t^{k_1}}{2}s_\theta(x_t, t, y)+\alpha_t^{\frac{k_1+k_2}{2}}z_t,~~~~z_t\sim\mathcal{N}(0,I)
\end{equation}
In practice, we choose $k_1=0.6$ and $k_2=0.4$ for faster convergence.
\section{Revised heat equation and target score computation}
\subsection{Revised heat equation}
\label{revisedheat}
It is able to solve the heat equation by utilizing the standard explicit method \cite{gerald2004applied}. Let's discretize $x,y,t$ as: $x_i=i \Delta x$, $y_j=j\Delta y$, and $k=n\Delta k$. For numerical stability, it is essential to satisfy $\Delta k\leq\frac{\Delta x^2}{4\alpha}$, under the assumption that $\Delta x=\Delta y$. By choosing $\Delta x=1$ to apply in $2D$ pixel space and $\Delta k=\frac{1}{4\alpha}$, the numerical approach can be denoted as:
\begin{equation}
    u^{n+1}_{i,j} = u^n_{i,j} + K_{i,j}\left (u^n_{i+1,j}+u^n_{i-1,j}+u^n_{i,j+1}+u^n_{i,j-1}-4 u^n_{i,j}  \right )
\end{equation}
The given equation has limitations: it is time-consuming since the heat value increases at a slow rate, leading to slow dispersion. Furthermore, empirical analysis suggests it is not appropriate for our method. Therefore, we have made modifications to the equation:
\begin{equation}
    u^{n+1}_{i,j} = u^n_{i,j} + K_{i,j}\left (u^n_{i+1,j}+u^n_{i-1,j}+u^n_{i,j+1}+u^n_{i,j-1}-V u^n_{i,j}  \right )
\end{equation}
where $V$ denotes the number of \textit{valid} neighboring states, specifically excluding insulators such as map edges and obstacles. 
After computing, the result is smoothed using a Gaussian filter to mitigate numerical inaccuracies. Subsequently, it is normalized by dividing by its sum, transforming it into a collision-avoiding diffusion kernel. We set $k$ in a range from 12.5 to 3612.5, with an exponential increase aligning with the progression of $t$. Here, we can interpret $p_{0t}$ as the heat distribution at time step $k$.\\
\subsection{Target score computation}
\label{targetscore}
We use a heat equation solver to calculate the target scores for our model, which incorporates an obstacle image mask, denoted as $y_{obs}$, and the positions of the goal, denoted as $x_0$. The obstacle image mask is a boolean tensor that has the same dimensions as the map image. It represents obstacles and map boundaries, which act as non-conductive insulators within the map. In the mask, these insulators are indicated by a value of 1. They influence the conductivity $K$ from the heat equation (Eq.\ref{heateq}).\\
The goal position is treated as the source of heat within this system. The heat equation solver then calculates the distribution of heat as it flows through the environment, taking into account the presence of obstacles, which the heat must navigate around, similar to how heat naturally avoids insulators.\\
The distribution is then normalized to represent a probability density function and is marginally adjusted by a small constant to ensure non-zero values before taking the logarithm. After logarithmic transformation, we obtain the gradient of this log-probability field through convolution with gradient kernels, creating the score field. The target score for any sampled state $x_t$ is then straightforwardly retrieved by substituting $x_t$ into this field. Our training objective is to have the model learn to approximate these target scores, thereby enabling the robot to infer collision-free paths toward its goal.

\newpage
\section{Baseline pseudo codes}
\label{pseudo}
\begin{algorithm}
\caption{Behavior Cloning model sampling}
\begin{algorithmic}[1]
\Require{input image $y$, stationary number $t$, sampling timesteps $N$}
  \State Initial states $x_0$
  \For{$i=0$ to $N$}
    \State $dx\leftarrow BC(y, t, x_i)$
    \State $x_{i+1}\leftarrow x_i+dx$ \Comment{Motion Planning}
  \EndFor
  \Return{$x_N$}
\end{algorithmic}
\end{algorithm}
\begin{algorithm}
\caption{Gaussian Diffusion sampling}
\begin{algorithmic}[1]
\Require{input image $y$, step size $\epsilon$}
  \State Initial states $x_T$
  \For{$t=T$ to $1$}    \Comment{Motion Planning with Each State Generated}
    \State $s_{\theta}\leftarrow Gaussian(y,t,x_t)$
    \State $x_{t-1}\leftarrow x_t+\frac{\epsilon}{2} s_\theta+\sqrt{\epsilon}z_t $ \Comment{$z_t\sim N(0,I)$}
  \EndFor
  \Return{$x_0$}
\end{algorithmic}
\end{algorithm}
\begin{algorithm}
\caption{Gaussian Diffusion + RRT* sampling}
\begin{algorithmic}[1]
\Require{input image $y$, step size $\epsilon$}
  \State Initial states $x_T$
  \For{$t=T$ to $1$}    \Comment{Goal Sampling}
    \State $s_{\theta}\leftarrow Gaussian(y,t,x_t)$
    \State $x_{t-1}\leftarrow x_t+\frac{\epsilon}{2} s_\theta+\sqrt{\epsilon}z_t $ \Comment{$z_t\sim N(0,I)$}
  \EndFor
  \State $x_0\leftarrow RRT^*(x_0,x_T)$    \Comment{Motion Planning toward the Generated Goal}\\
  \Return{$x_0$}
\end{algorithmic}
\end{algorithm}
\end{document}